\documentclass[sigconf]{acmart}
\usepackage{amsmath,amsfonts}
\usepackage{graphicx}
\usepackage{adjustbox}
\usepackage{float}
\usepackage{textcomp}
\usepackage{xcolor}
\usepackage{hhline}
\usepackage{algorithmicx}
\usepackage{algorithm}
\usepackage{algpseudocode}
\usepackage{multirow}
\usepackage{fancyvrb}
\usepackage{enumitem}
\usepackage{blindtext}
\usepackage{hyperref}
\usepackage{subcaption}

\setlist[itemize]{topsep=3pt}

\AtBeginDocument{%
  }

\setcopyright{acmlicensed}
\copyrightyear{2018}
\acmYear{2018}
\acmDOI{XXXXXXX.XXXXXXX}

\begin{document}

\title{TrendGen: An Outfit Recommendation and Display System}

\author{Theodoros Koukopoulos}
\email{thodoris.koukopoulos@sabinodb.com}
\affiliation{%
  \institution{SabinoDB}
  \city{Athens}
  \country{Greece}}

\author{Dimos Klimenof}
\email{dimosthenis.klimenof@sabinodb.com}
\affiliation{%
  \institution{SabinoDB}
  \city{Athens}
  \country{Greece}}
  
\author{Ioannis Xarchakos}
\email{ioannis.xarchakos@sabinodb.com}
\additionalaffiliation{University of Toronto}
\affiliation{%
  \institution{SabinoDB}
  \city{Athens}
  \country{Greece}}

\begin{abstract}

Recent advances in Computer Vision have significantly improved image understanding and generation, revolutionizing the fashion industry. However, challenges such as inconsistent lighting, non-ideal garment angles, complex backgrounds, and occlusions in raw images hinder their full potential. Overcoming these obstacles is crucial for developing robust fashion AI systems capable of real-world applications.

In this paper, we introduce TrendGen, a Fashion AI system designed to enhance online shopping with intelligent outfit recommendations. Deployed on a major e-commerce platform, TrendGen leverages cloth images and product attributes to generate trend-aligned, cohesive outfit suggestions. Additionally, it employs Generative AI to transform raw images into high-quality lay-down views, offering a clear and structured presentation of garments.

Our evaluation on production data demonstrates TrendGen’s consistent high-quality outfits and lay-down images, marking a significant advancement in AI-driven solutions for fashion retail.

\end{abstract}

\maketitle
\section{Introduction}
\label{sec:intro}

The fashion industry is a vital pillar of the global economy, generating over \$1.7 trillion annually. Beyond its economic contributions, it shapes cultural trends and societal norms on a global scale. Advanced technologies like natural language processing, computer vision, and machine learning are transforming operations by improving efficiency, optimizing supply chains, and reducing costs. These innovations enable precise demand forecasting, better inventory management, and reduced waste. Computer vision further enhances user experiences through personalized shopping, virtual try-ons, and smarter product recommendations, driving customer engagement and revenue growth.

Recent advances in Deep Learning (DL) have revolutionized computer vision, excelling in image classification, object detection, segmentation, and generation. In fashion, DL enhances efficiency, creativity, and customer engagement \cite{ding2023computational, cheng2021fashion}. For example, product attribution automates the labeling of style, pattern, and color, while image retrieval enables seamless discovery of visually similar products. Outfit recommendations personalize shopping by analyzing user preferences, and virtual try-ons simulate garment fit, transforming the online shopping experience.

Previous studies on fashion computer vision often focus on isolated applications using curated datasets under controlled conditions, failing to capture real-world complexities. These works prioritize high accuracy within a narrow scope but overlook challenges like missing data, diverse inputs, and system integration. In contrast, production environments face a variety of issues, including inconsistent image quality, variable lighting, diverse angles, and complex backgrounds, requiring models to be robust and adaptable. While isolated research offers valuable insights, it does not fully prepare systems for real-world deployment, where reliability, scalability, and adaptability are essential.

In this paper, we introduce TrendGen, a Fashion AI system that is currently deployed on a major online shopping store, designed to enhance the user's online shopping experience by generating complete outfits for every available product. TrendGen’s outfit recommendations simplify decision-making, boost consumer confidence, and increase conversion rates. By suggesting complete outfits, the system encourages additional purchases and fosters brand loyalty, contributing to a 5\% increase in product sales since its deployment. As personalization and convenience become crucial in the digital marketplace, TrendGen’s smart outfit recommendations serve as vital tools for maintaining competitive advantage and driving growth.

\begin{figure}
\centering
\begin{subfigure}{0.35\linewidth}
    \centering
    \includegraphics[width=\linewidth]{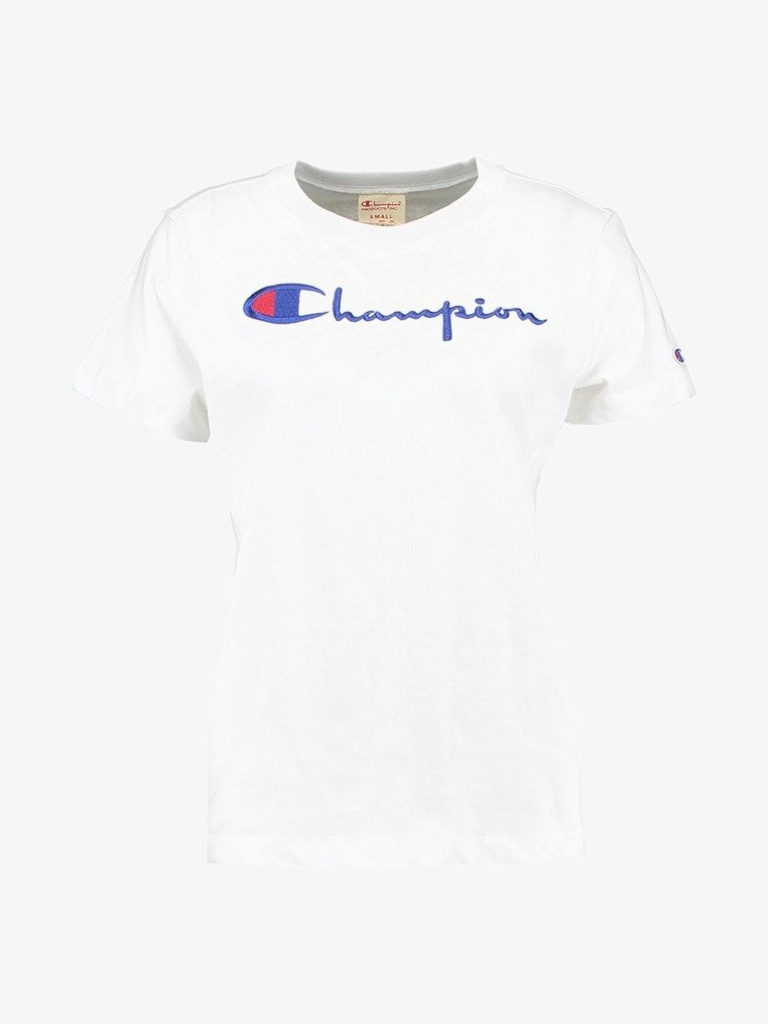}
    \caption{Tiled cloth}
\end{subfigure}\hfill
\begin{subfigure}{0.35\linewidth}
    \centering
    \includegraphics[width=\linewidth]{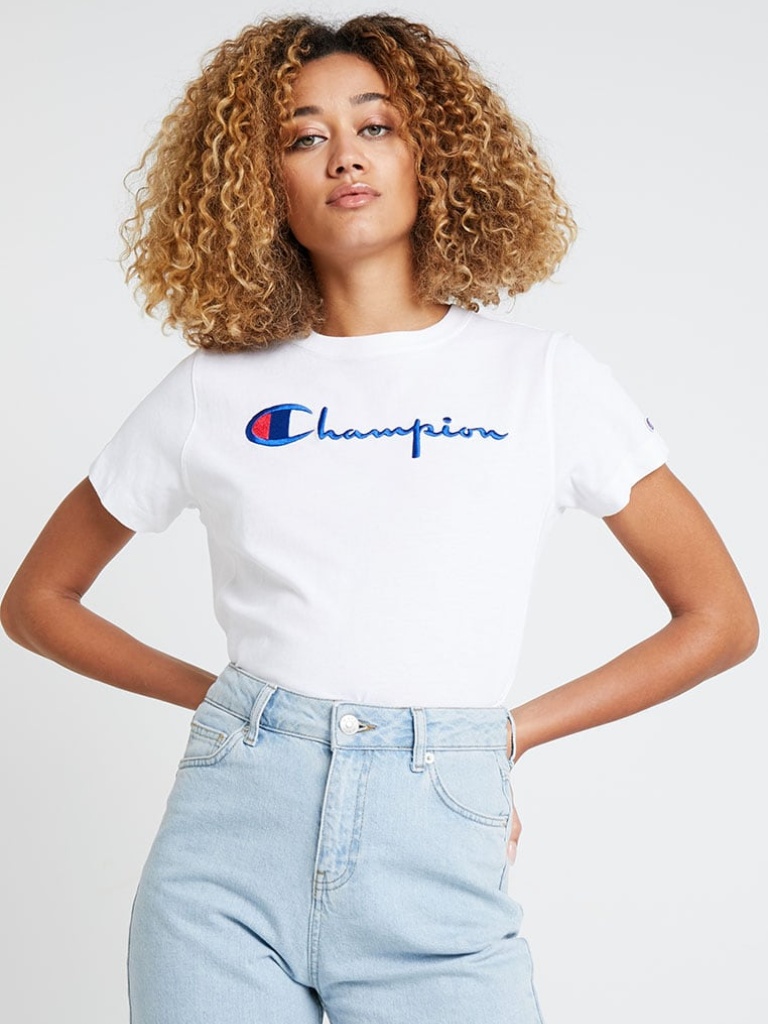}
    \caption{Human worn cloth}
\end{subfigure}
\caption{Different garment image views}
\label{fig:garment}
\end{figure}

Previous outfit recommendation methods \cite{gomez2017selfsupervisedlearningvisualfeatures, 6a9c1801d0de4c39965371fe0596b536, Han_2017, he2016learningcompatibilitycategoriesheterogeneous, veit2015learningvisualclothingstyle, Sarkar_2023_WACV} rely on lay-down product images (Figure \ref{fig:garment}a) and text descriptions, as input to their recommendation engines. Utilizing raw text, however, results in suboptimal performance since product descriptions contain non-essential details that do not help on making better outfit recommendations. To this end, we propose to transform raw text, to meaningful attributes that help the outfit recommendation engine make more intelligent outfit combinations.

An additional challenge we face is that our products are typically introduced with only the human-worn garment image (Figure \ref{fig:garment}b), along with a product title and description. However, raw images frequently contain distracting elements like human models and cluttered backgrounds, while garments are often captured at poor angles, under suboptimal lighting, or partially obscured by features like hair or limbs. To address this challenge, TrendGen introduces an innovative image generation model, which converts human-worn garment images into lay-flat representations (Figure \ref{fig:garment}a). This virtual try-off process \cite{velioglu2024tryoffdiffvirtualtryoffhighfidelitygarment, xarchakos2024tryoffanyonetiledclothgeneration, shen2024igrimprovingdiffusionmodel} removes distractions, ensuring clean, consistent inputs for outfit recommendations. Beyond enhancing recommendation accuracy, these lay-down images are displayed online, offering shoppers a clear, structured view of each product of the outfit.

\begin{figure}
\centering
\begin{subfigure}{0.35\linewidth}
    \centering
    \includegraphics[width=\linewidth]{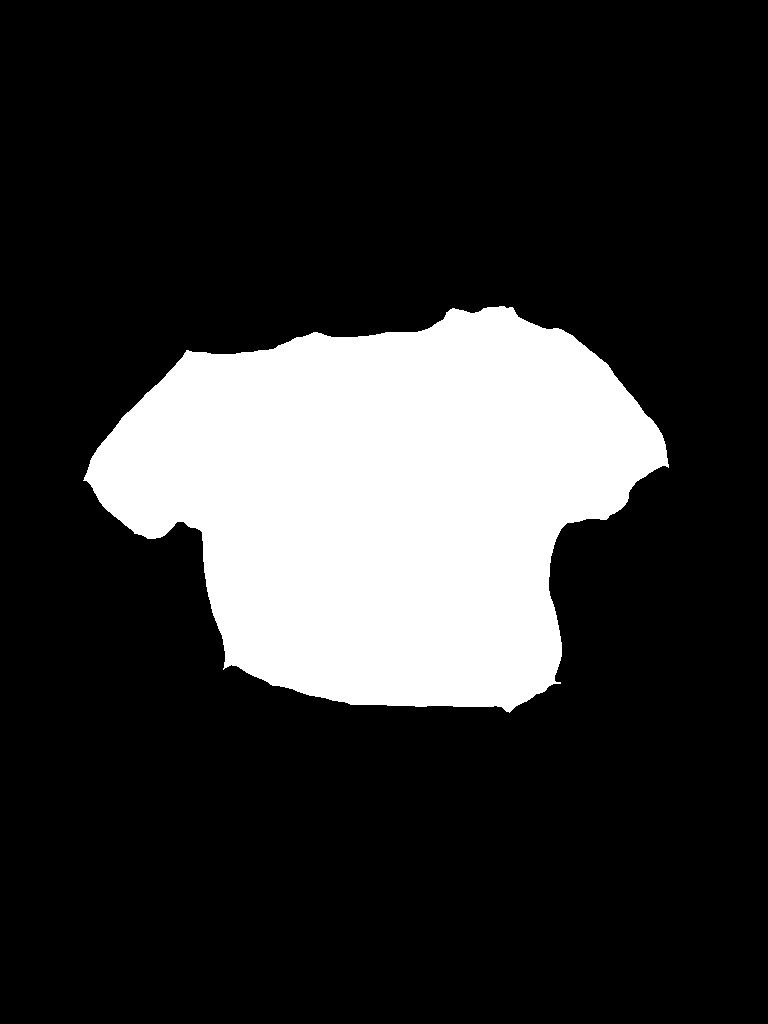}
    \caption{Binary cloth mask}
\end{subfigure}\hfill
\begin{subfigure}{0.35\linewidth}
    \centering
    \includegraphics[width=\linewidth]{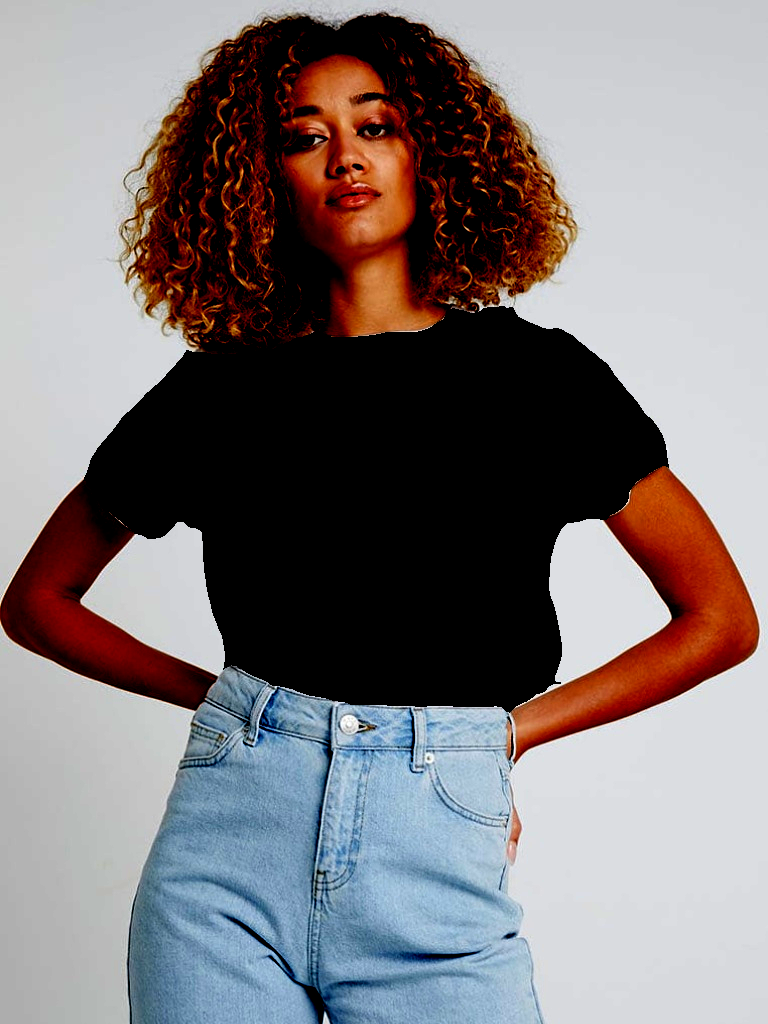}
    \caption{Masked cloth}
\end{subfigure}
\caption{Cloth masking}
\label{fig:mask}
\end{figure}

In particular, TrendGen first extracts product attributes from the human-worn garment image and product title. In Section \ref{sec:attribution}, we present our product attribution engine, which is responsible for generating a variety of product attributes, such as product category, color, hem type, sleeve length, etc. We utilize  Fashion-CLIP \cite{Chia2022} to generate embeddings from the image and text. Then, we train a classifier to predict each attribute of interest.

Following the attribution of a product, we invoke our virtual try-off model of Section \ref{sec:laydowns}, to generate high-quality and high-resolution lay-down garment images from images of garments worn by humans. We utilize a Latent Diffusion Model (LDM) \cite{rombach2022highresolutionimagesynthesislatent} pre-trained on the inpainting variant of StableDiffusion v1.5 \cite{rombach2022highresolutionimagesynthesislatent}. We then fine-tune the U-Net using the human-worn garment images (Figure \ref{fig:garment}b) as input and their corresponding tiled garment images (Figure \ref{fig:garment}a) as targets, from the VITON-HD dataset \cite{DBLP:journals/corr/abs-2103-16874}. To optimize the model performance for our specific application, we further fine-tune the model on a limited set of paired tiled and human-worn garment images of our product data.

After generating lay-flat images, we use the outfit recommendation engine, as outlined in Section \ref{sec:outfits}, to create complete, cohesive outfits from our product data. To achieve this, we leverage Fashion-CLIP \cite{Chia2022}, which combines the tiled garment images with their corresponding product attributes to generate robust feature representations for each item. These representations are mapped into refined feature spaces, where items from different categories (e.g., Tops and Bottoms) are positioned closer together if they exhibit aesthetic compatibility. This process is driven by a contrastive learning approach, which uses positive and negative examples to optimize the alignment of compatible products across categories. While our dataset includes only positive examples, we address this limitation by incorporating domain expert knowledge and leveraging product attributes to select meaningful negative samples for model training. This ensures that the system learns nuanced distinctions between compatible and incompatible items, enhancing its ability to generate visually appealing and contextually appropriate outfits.

Unlike previous approaches \cite{gomez2017selfsupervisedlearningvisualfeatures, 6a9c1801d0de4c39965371fe0596b536, Han_2017, he2016learningcompatibilitycategoriesheterogeneous, veit2015learningvisualclothingstyle, Sarkar_2023_WACV}, which generate outfits in isolation based solely on stylistic compatibility, our method ensures diversity by creating three distinct outfits for each product. We propose a simple yet effective algorithm that balances aesthetic similarity with item frequency in prior recommendations, preventing repetition and enhancing outfit variety to better align with diverse consumer preferences.

Specifically, we make the following contributions:
\begin{itemize}

    \item We propose a simple and precise product attribution system for fashion items 

    \item We propose a state-of-the-art Latent Diffusion Model (LDM) for the Virtual Try-Off task

    \item We propose an outfit recommendation engine that produces high-quality, diverse outfits by ranking products based on aesthetic similarity as well as repetitiveness 

    \item We conduct experiments on our product data, analyzing each proposed component of our system in isolation for its suitability and effectiveness for the associated task as well as the end-to-end system performance
    
\end{itemize}

This paper is organized as follows: In Section \ref{sec:related} we review the related work. Section \ref{sec:attribution} provides details about the production attribution engine, while Section \ref{sec:laydowns} introduces our novel diffusion-based lay-down cloth generation model. Section \ref{sec:outfits} demonstrates our outfit recommendation engine. Section \ref{sec:experimental} presents our experimental evaluation. Section \ref{sec:conc} concludes the paper by discussing avenues to future work in this area.

\begin{figure*}[h!]
\centering
\includegraphics[width=0.9\textwidth]{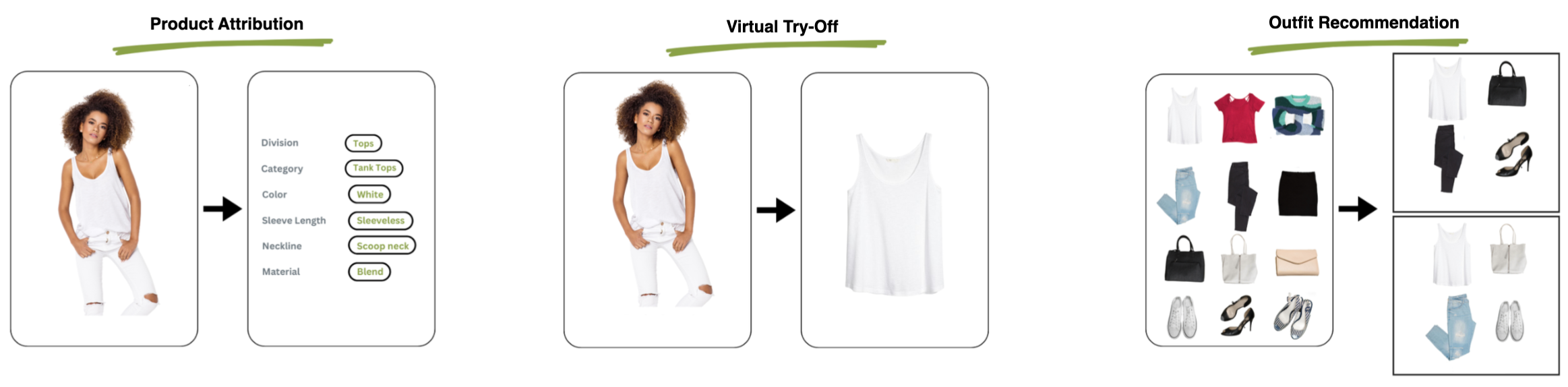}
\caption{TrendGen system components}
\label{fig:overall}
\end{figure*}

\section{Related Work}
\label{sec:related}

\subsection{Multi-modal classification}
Multi-modal classification, which integrates visual and textual data, has garnered significant attention in recent years due to its potential to enhance model performance by leveraging complementary information from multiple modalities. Early approaches, such as Gallo et al. \cite{8615789}, proposed embedding encoded text onto images to create information-enriched inputs, enabling Convolutional Neural Networks (CNNs) to learn joint representations for improved classification accuracy. Subsequent research introduced more sophisticated architectures to capture complex relationships between modalities. Mafla et al. \cite{mafla2020multi} developed a Multi-Modal Reasoning Graph that employs Graph Convolutional Networks to model interactions between visual elements and scene text, enhancing fine-grained image classification and retrieval tasks. The advent of large-scale pre-trained models further advanced the field. Kaul et al. \cite{Kaul2023} utilized Contrastive Language-Image Pre-training (CLIP) \cite{Radford2021LearningTV} to perform open-vocabulary object detection, demonstrating the efficacy of aligning visual and textual representations in a shared embedding space for detecting objects beyond the set of categories seen during training. More recently, Guo \cite{guo2024multimodalmultilabelclassificationclip} explored multimodal multilabel classification by fine-tuning CLIP with various classification heads, fusion methods, and loss functions, achieving significant improvements in performance. Such methods can be crucial for fashion garment classification, where garments often belong to multiple overlapping categories (e.g., “sporty yet formal”) that require a multi-label classification approach.

\subsection{Virtual Try-Off}
The rapid advancements in latent diffusion models (LDMs), spearheaded by Stable Diffusion \cite{rombach2022highresolutionimagesynthesislatent}, have significantly propelled progress in text-to-image generation \cite{esser2024scalingrectifiedflowtransformers, podell2023sdxlimprovinglatentdiffusion, rombach2022highresolutionimagesynthesislatent}. To enhance control over generated images, several universal controllers \cite{Mou_Wang_Xie_Wu_Zhang_Qi_Shan_2024, ye2023ipadaptertextcompatibleimage, zhang2023addingconditionalcontroltexttoimage} have been proposed. For instance, IP-Adapter \cite{ye2023ipadaptertextcompatibleimage} enables LDMs to incorporate input image prompts for generating outputs that retain specific features, while attribute-specific ControlNet \cite{zhang2023addingconditionalcontroltexttoimage} aligns outputs with defined poses, outlines, depths, and other attributes. Similarly, T2I-Adapter \cite{Mou_Wang_Xie_Wu_Zhang_Qi_Shan_2024} introduces a unified model to accept various attributes. These controllers can be easily integrated into most
downstream tasks \cite{choi2024improvingdiffusionmodelsauthentic, xu2023magicanimatetemporallyconsistenthuman}. However, specialized tasks often demand tailored approaches. For example, pose-guided human generation \cite{hu2024animateanyoneconsistentcontrollable, lu2024coarsetofinelatentdiffusionposeguided, xu2023magicanimatetemporallyconsistenthuman} requires advanced controls to maintain identity consistency during pose transformations, while VITON \cite{choi2024improvingdiffusionmodelsauthentic, kim2024stableviton, xu2024ootdiffusionoutfittingfusionbased} relies on feature extractors to isolate complete garment features against clean backgrounds. The goal of garment restoration is to reconstruct standard garments from person images. TileGAN \cite{article_TileGAN} was the first to introduce a GAN-based model \cite{NIPS2014_5ca3e9b1, 8100115} for this purpose, using category guidance to generate coarse garments in the first stage, followed by refinement in the second. Recent methods like TryOffDiff \cite{velioglu2024tryoffdiffvirtualtryoffhighfidelitygarment}, TryOffAnyone \cite{xarchakos2024tryoffanyonetiledclothgeneration}, IGR \cite{shen2024igrimprovingdiffusionmodel}, TEMU-VTOFF \cite{lobba2025inversevirtualtryongenerating}, MGT \cite{velioglu2025mgtextendingvirtualtryoff} and RAGDiffusion \cite{tan2024ragdiffusionfaithfulclothgeneration} leverage pretrained LDMs \cite{rombach2022highresolutionimagesynthesislatent} to achieve superior generative performance.

\subsection{Outfit Recommendation}
Fashion recommendation systems have gained significant attention in recent years due to their potential to revolutionize the shopping experience by providing personalized suggestions. A popular approach is to use embedding-based approaches that offer a framework for learning complex relationships by leveraging positive and negative sample pairs. These methods are often implemented using siamese networks \cite{Siamese} or trained with triplet loss functions \cite{NIPS2003_d3b1fb02}, demonstrating remarkable performance on challenging tasks like face verification \cite{FaceNet}. This capability is particularly appealing for fashion-related tasks, which often involve learning intricate, hard-to-define relationships between items \cite{gomez2017selfsupervisedlearningvisualfeatures, 6a9c1801d0de4c39965371fe0596b536, Han_2017, he2016learningcompatibilitycategoriesheterogeneous, veit2015learningvisualclothingstyle}. For instance, Veit et al. \cite{veit2015learningvisualclothingstyle} achieved large-scale similarity and compatibility matching for clothing images. However, their approach does not differentiate items based on type (e.g. tops, shoes, or scarves) within the embeddings. Han et al. \cite{Han_2017} addressed this by using an LSTM model to process the visual representation of each item in an outfit, enabling holistic outfit reasoning.

Some methods focus on learning multiple embeddings \cite{10.1145/2766959, inproceedingsChen, 6a9c1801d0de4c39965371fe0596b536, song2017learningunifiedembeddingapparel, veit2017conditionalsimilaritynetworks}, often assuming separation between item types (e.g. comparing bags exclusively with other bags) or employing knowledge graph representations \cite{xiao2017sspsemanticspaceprojection}. Multi-modal embeddings have also been shown to uncover novel feature structures \cite{gomez2017selfsupervisedlearningvisualfeatures, Han_2017, 35a9d959d8d24c2daa1ed5b6a21c078d, 8099810_Salvador}, which this work also incorporates in Section \ref{sec:attribution}. Nevertheless, training embedding networks can be challenging, as arbitrary triplet selection often results in weak constraints \cite{wu2018samplingmattersdeepembedding, zhuang2016fasttrainingtripletbaseddeep}. 

OutfitTransformer \cite{Sarkar_2023_WACV} eliminates the need for triplet selection by providing one or more items of the outfit as input to a transformer which returns compatible items of other categories. This work is the current state-of-the-art in outfit recommendation and we compare against in Section \ref{sec:experimental}.


\section{Product Attribution Engine}
\label{sec:attribution}
\begin{figure*}
\centering
\includegraphics[width=0.76\textwidth]{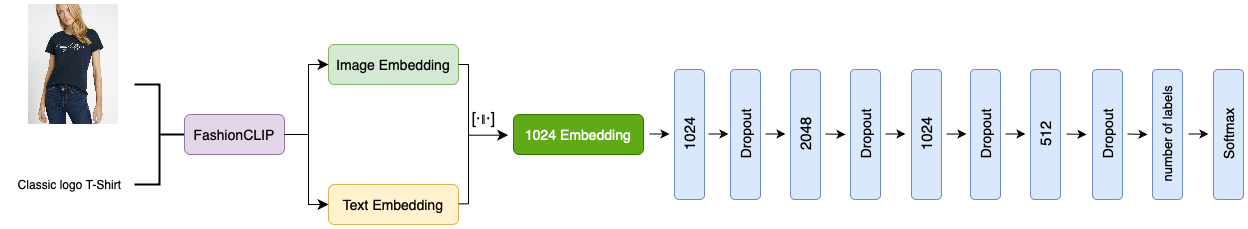}
\caption{Attribution Classifier Architecture}
\label{fig:classifier}
\end{figure*}

\subsection{Feature Extraction}
 
Our proposed system, TrendGen, incorporates an advanced attribution mechanism that combines the raw image of a product with its stylist-provided title to generate rich and informative embeddings. At the core of this mechanism lies FashionCLIP \cite{Chia2022}, a domain-adapted version of the CLIP \cite{Radford2021LearningTV} model tailored specifically for the fashion industry. FashionCLIP builds on the foundational architecture of CLIP by leveraging contrastive language–image pretraining, optimized to learn representations of general fashion concepts. This adaptation allows it to capture fine-grained relationships between visual and textual data, ensuring a robust and domain-relevant shared embedding space. The embeddings produced by FashionCLIP serve as the input to the final classifier, enabling precise and detailed product categorization. By utilizing FashionCLIP, the system benefits from its ability to integrate complementary information from both image and text modalities, which is critical in the fashion domain where context and aesthetics play a significant role. This multi-modal approach significantly outperforms traditional independent models, such as convolutional neural networks (CNNs) for image features and natural language processing (NLP) networks for textual features.

\subsection{Attribute Classifier}

Our system’s final classification module leverages FashionCLIP embeddings for precise and efficient product categorization. To ensure simplicity and scalability, we conducted extensive experiments to identify the optimal neural network architecture, ultimately adopting the design shown in Figure \ref{fig:classifier}.

This architecture features fully connected (dense) layers that progressively refine embeddings into classification-ready features. It maintains a consistent structure across all attribution levels, with only the output layer’s neuron count varying based on category granularity. Each dense layer incorporates the Parametric Rectified Linear Unit (PReLU) \cite{7410480}, which dynamically learns rectifier parameters, enhancing adaptability to complex data distributions.

To prevent overfitting, we integrate Dropout layers after each dense layer, applying a 0.5 dropout probability. This regularization technique ensures robust generalization, allowing the model to handle diverse and challenging inputs effectively.

\section{Virtual Try-Off}
\label{sec:laydowns}

An essential component of the TrendGen framework is the virtual try-off module, which transforms an image of a garment worn by a person (Figure \ref{fig:garment}b) into a lay-down view of the garment (Figure \ref{fig:garment}a), to provide them as input to the outfit recommender as well as to enhance the display of individual clothing items.

\subsection{Network Architecture}
To generate lay-down images of the garments, we utilize the inpainting variant of Stable Diffusion 1.5 \cite{rombach2022highresolutionimagesynthesislatent}, a Latent Diffusion Model (LDM) trained on LAION dataset \cite{schuhmann2022laion5bopenlargescaledataset}. LDMs map high-dimensional image inputs to a lower-dimensional latent space using a pre-trained Variational Autoencoder (VAE)  \cite{kingma2022autoencodingvariationalbayes}, reducing computational cost while maintaining high-quality outputs.

An LDM comprises of a denoising UNet $E_{\theta}(\cdot, t)$ and a VAE with an encoder $\epsilon$ and decoder $D$. The training minimizes:
\[
\mathcal{L}_{\text{LDM}} := \mathbb{E}_{\epsilon(x), \epsilon \sim \mathcal{N}(0, 1), t} 
\left[\| \epsilon - \epsilon_{\theta}(z_t, t) \|_2^2\right]     
\]
where \( z_t \) represents the noisy latent encoding at timestep \( t \). The forward process adds Gaussian noise \( \mathcal{N}(0, 1) \) to the encoded input, while the reverse process iteratively denoises \( z_t \) to reconstruct \( z_0 \). Finally, \( D \) decodes \( z_0 \) back into the image domain.

Unlike Stable Diffusion \cite{rombach2022highresolutionimagesynthesislatent}, which relies on text encoders such as CLIP \cite{Radford2021LearningTV} to condition the generation process, our proposed architecture eliminates the need for textual descriptions by leveraging a cloth mask (Figure \ref{fig:mask}b) that covers the target garment. This mask-based guidance significantly improves the generation process in two key aspects; it enhances image quality by providing explicit localization of the target garment, reducing ambiguities introduced by textual prompts, while it streamlines the training pipeline by removing the computational overhead associated with text encoders, thereby improving training efficiency.

\begin{figure*}[h!]
\centering
\includegraphics[height=0.41\linewidth]{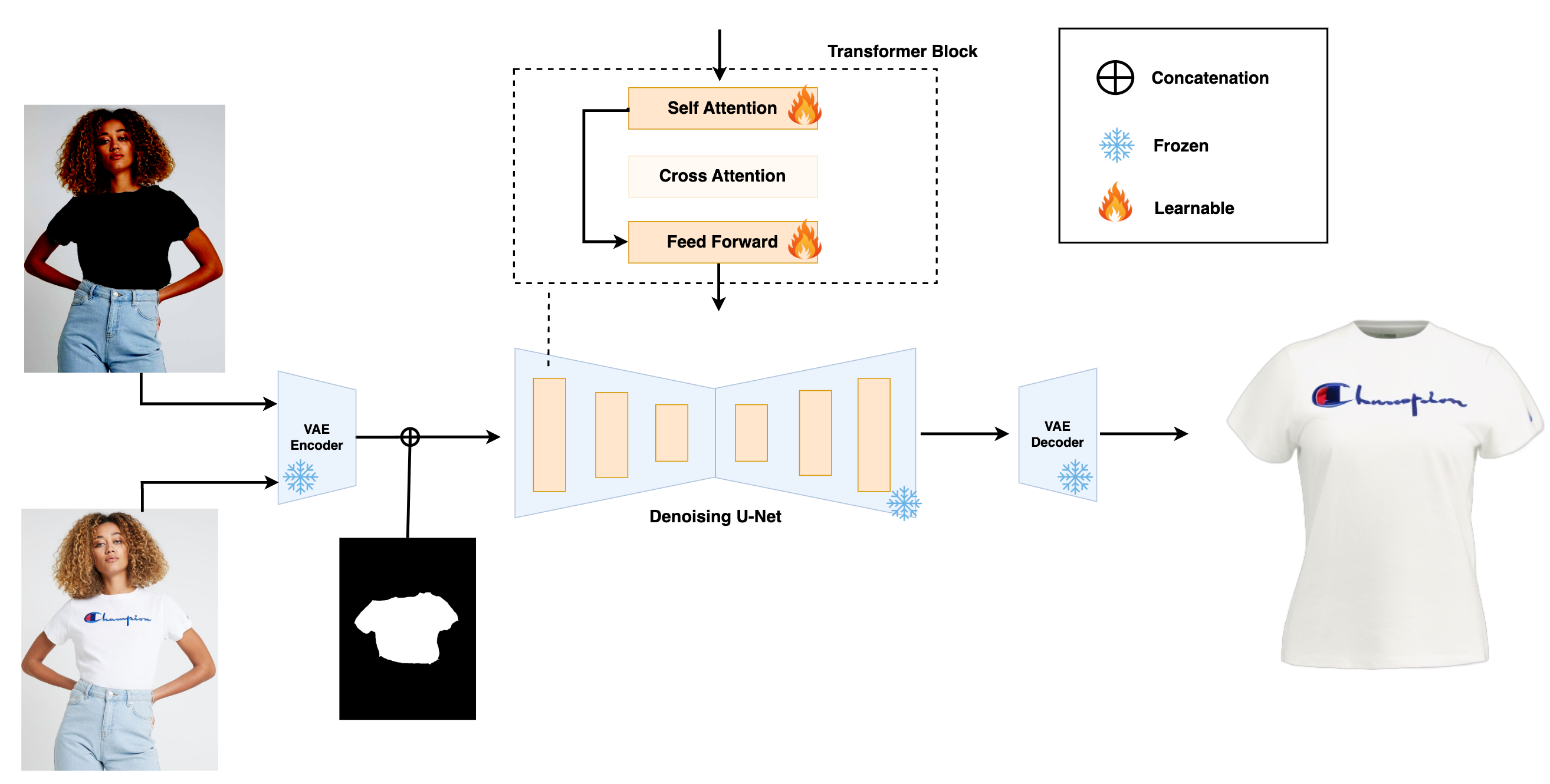}
\caption{Virtual Try-Off Network Architecture}
\label{fig:vitoff}
\end{figure*}

The proposed architecture consists of two primary modules: a pre-trained Variational Autoencoder (VAE) and a denoising U-Net. The VAE operates as a latent space encoder-decoder, compressing the input image into a lower-dimensional representation while preserving essential garment features. The U-Net, conditioned on the dressed person's cloth (Figure \ref{fig:garment}b) and the cloth mask (Figure \ref{fig:mask}a) latent representations, performs iterative denoising to generate the target lay-down cloth image (Figure \ref{fig:garment}a). The proposed model architecture is presented in Figure \ref{fig:vitoff}.

\subsection{Network Training}
Our experiments showed that fine-tuning only the transformer blocks of the U-Net delivers the best performance, outperforming both attention-only and full-network training. Fine-tuning only the attention layers proved inadequate for the complex task of generating virtual try-offs, as it failed to capture the intricate structural and textural garment details. Conversely, while fine-tuning the entire architecture produced comparable results, it demands substantially higher computational resources, making it less practical.

To fine-tune the proposed architecture for the virtual try-off task, we utilize pairs of lay-down images (Figure \ref{fig:garment}a) and human worn cloth images (Figure \ref{fig:garment}b). Furthermore, we extract a binary cloth mask (Figure \ref{fig:mask}a) and we additionally generate the original raw image with the garment masked (Figure \ref{fig:mask}b). The cloth mask serves as explicit guidance for the generation process, indicating the precise garment to be reconstructed in a lay-flat configuration. This approach addresses challenges posed by complex scenes in the input image, such as the presence of multiple garments or occlusions, ensuring accurate extraction and generation of the desired clothing item. Full implementation details are presented elsewhere \cite{xarchakos2024tryoffanyonetiledclothgeneration}.

\section{Outfit Recommendation Engine}
At the core of our system is the outfit recommendation module, designed to generate highly stylized, cohesive outfit suggestions that are both aesthetically pleasing and drive purchases. Built on a robust TripleNet framework with a contrastive loss function, it learns stylistic compatibility embeddings between fashion items. This approach enables well-curated, harmonious outfits by integrating visual and textual data through a meticulous pipeline, ensuring recommendations are both fashion-forward and commercially viable.

\label{sec:outfits}

\subsection{Embedding Generation}
The system starts with images from the virtual try-off module, depicting garments without a wearer. Along with product attributes (color, division, category), these images are processed by FashionCLIP \cite{Chia2022} to generate 512-dimensional image and text embeddings. These embeddings are concatenated into a 1024-dimensional multimodal representation for training.

\[
\mathcal{E}(\mathbf{I}, \mathbf{P}) = [\mathcal{E}_{\text{img}}(\mathbf{I}) \, \Vert \, \mathcal{E}_{\text{text}}(\mathbf{P})]
\]

Where:
\begin{itemize}
    \item \(\mathcal{E}_{\text{img}}(\mathbf{I}) \in \mathbb{R}^{512}\): Image embedding from FashionCLIP, capturing visual features from the lay-down image \(\mathbf{I}\).
    \item \(\mathcal{E}_{\text{text}}(\mathbf{P}) \in \mathbb{R}^{512}\): Text embedding generated by FashionCLIP, using product attributes  \(\mathbf{P}\) as input.
    \item \(\mathbf{P}\): Attribute vector describing the garment, comprises of Division (e.g., Tops, Bottoms, Accessories), Category (e.g., T-shirts, Pants, Bags), and Color
    
    \item \([\cdot \Vert \cdot]\): Concatenating the two 512-dimensional embeddings \(\mathcal{E}_{\text{img}}(\mathbf{I}), \mathcal{E}_{\text{text}}(\mathbf{P})\) into a 1024-dimensional representation.
\end{itemize}

By integrating both visual and textual features, this multimodal embedding enables the system to evaluate visual compatibility alongside semantic garment attributes, resulting in stylistically cohesive and domain-appropriate recommendations.

\subsection{TripletNet}
TripleNet processes three inputs; anchor, positive, and negative embeddings, through a shared neural network, ensuring consistent transformations. The architecture is as follows:

\begin{itemize}
    \item \textbf{Input Layer:} A 1024-dimensional embedding for each of the anchor, positive, and negative samples.
    \item \textbf{Shared Dense Layer:} A fully connected layer reduces the dimensionality of the embeddings from 1024 to 512.
    \item \textbf{Output Layer:} A 512-dimensional embedding for each input (anchor, positive, and negative).
\end{itemize}

\subsubsection{Loss Function}
The proposed recommendation system utilizes a Triplet Loss function to optimize the embedding space. The Triplet Loss encourages the embedding of an anchor sample to be closer to a positive sample (stylistically compatible item) than to a negative sample (incompatible item) by at least a margin \(m > 0\). The loss function is formally defined as:

\[
\mathcal{L} = \sum_{i=1}^{N} \max\left(0, \|\mathbf{f}_{a}^{(i)} - \mathbf{f}_{p}^{(i)}\|_2^2 - \|\mathbf{f}_{a}^{(i)} - \mathbf{f}_{n}^{(i)}\|_2^2 + m\right)
\]

Where:
\begin{itemize}
    \item \(\mathbf{f}_{a}\), \(\mathbf{f}_{p}\), and \(\mathbf{f}_{n}\) are the embeddings of the anchor, positive, and negative samples respectively.
    \item \(m\) is the margin that enforces a minimum distance between positive and negative pairs relative to the anchor.
    \item \(\|\cdot\|_2^2\) represents the squared Euclidean distance.
\end{itemize}

The goal of minimizing this loss function is to ensure that positive pairs are closer in the embedding space than negative pairs by at least the specified margin.

\subsection{Triplets Construction}
Training the TripleNet framework involves constructing triplets: an anchor, a positive, and a negative sample, where the anchor is the product to be recommended, the positive sample is a stylistically compatible item from a different division, while the negative sample is a stylistically mismatched item from the same division as the positive. The triplet design encourages the network to bring compatible items closer in the embedding space while pushing incompatible items further apart. The selection of positives and negatives follows some specific rules:
\begin{enumerate}
\item \textbf{Bottoms, Footwear, Outerwear Anchors:} Positive is always from the \textit{Tops} division, as matching these with a top garment is stylistically critical.
\item \textbf{Accessories Anchors:} Positive is randomly selected from \textit{Tops} or \textit{Bottoms}, ensuring compatibility with both divisions.
\item \textbf{Tops Anchors:} Positive can be from any other division for flexible styling options.
\end{enumerate}

Separate TripleNet models are trained for each division pairing (e.g., Tops-Bottoms, Bottoms-Tops, Tops-Footwear) to capture specific inter-division compatibility.

\subsubsection{Negative Sample Mining}
Our negative sample mining strategy incorporates expert-driven fashion design principles to ensure the construction of semantically meaningful and stylistically relevant triplets. By leveraging expert guidelines, the system avoids pairing garments that violate fundamental styling rules, thereby preventing unrealistic or undesirable recommendations. For instance, a key rule applied is the exclusion of outfits containing multiple multicolor garments, as such combinations are considered visually discordant. To enforce this constraint, triplets are deliberately constructed with a multicolor anchor paired with a multicolor negative sample, enabling the model to learn that such combinations should never be recommended. Beyond this, additional proprietary styling rules provided by fashion designers further refine the negative sample selection process enhancing the model’s ability to generate fashion recommendations that align with expert styling conventions and user expectations.

\subsection{Outfit Recommendation}

A significant limitation of existing outfit recommendation methods \cite{gomez2017selfsupervisedlearningvisualfeatures, 6a9c1801d0de4c39965371fe0596b536, Han_2017, he2016learningcompatibilitycategoriesheterogeneous, veit2015learningvisualclothingstyle} is their tendency to suggest aesthetically pleasing items without considering prior recommendations. Most algorithms focus solely on maximizing visual appeal, disregarding the repetitiveness of selected products across multiple outfit suggestions. Given the extensive range of products available for outfit generation, it is essential to account for the frequency with which a product appears in the recommended outfits. To enhance both diversity and aesthetic coherence, we propose STYLERANK (Algorithm \ref{alg1}), an effective yet simple algorithm that adjusts a product’s ranking based on its occurrence in previously generated outfits.

Algorithm \ref{alg1} evaluates a product’s outfit compatibility and historical appearance frequency. Using k-Nearest Neighbors (kNN) on TripletNet embeddings, it ranks the Top-K most aesthetically compatible items relative to a target product (anchor), from 1 (most compatible) to K (least). K is capped at 100 to focus on relevant matches, as items beyond this range show weak compatibility based on qualitative observations. This threshold, though not empirically validated, balances relevance and efficiency.

\begin{center}
$
\text{D}_{\text{bottom}} = \text{k-NN}(\mathcal{E}(\mathbf{I}_{\text{top}}, \mathbf{P}_{\text{top}}), \mathcal{P}_{\text{bottom}}, k=100)$

    $
\text{D}_{\text{accessories}} = \text{k-NN}(\mathcal{E}(\mathbf{I}_{\text{top}}, \mathbf{P}_{\text{top}}), \mathcal{P}_{\text{accessory}}, k=100)
$

\end{center}
Where:

\begin{itemize}
    \item \(\text{k-NN}(\cdot)\): The k-Nearest Neighbors algorithm, used to retrieve the \(k\) most relevant garments from a pool based on the distance in the embedding space.
    \item \(\mathcal{E}(\mathbf{I}_{\text{top}}, \mathbf{P}_{\text{top}})\): The embedding generated by the FashionCLIP model, which combines visual and textual information.
    \begin{itemize}
        \item \(\mathbf{I}_{\text{top}}\): The lay-down image of the "top" garment.
        \item \(\mathbf{P}_{\text{top}}\): The attribute vector of the "top" garment, including division, category, and color.
    \end{itemize}
    \item \(\mathcal{P}_{\text{bottom}}\): The pool of bottoms (e.g., pants, skirts)
    \item \(\mathcal{P}_{\text{accessories}}\): The pool of accessories (e.g., jewelry, bags)
    \item \(k\): The number of garments matched to the anchor garment.
\end{itemize}

Next, we rerank the Top-K products using an appearance count table $\mathcal{A}_p$, assigning the highest rank (100) to the product with the highest frequency in previous outfits. A control parameter $\lambda$ modulates the influence of appearance frequency on the final score. We set $\lambda = 1$ to balance compatibility and repeatability, as justified through ablation in Section~\ref{appendix_stylerank}. The final score for each product is the sum of its compatibility and appearance ranks, and the product with the lowest combined score is selected. The algorithm then constructs the outfit by sequentially selecting the most compatible items across categories—e.g., bottoms ($\text{match}_{\text{top-bottom}}$) and accessories ($\text{match}_{\text{top-accessory}}$)—repeating this process for all required garment types. The final outfit consists of the top-ranked item from each category.

\begin{center}
\[
\text{match}_{\text{top-bottom}} =
\mathrm{STYLERANK}(\mathcal{P}_{\text{bottom}}, \mathcal{D}_{\text{bottom}}, \mathcal{A}_{p})
\]

\[
\text{match}_{\text{top-accessory}} =
\mathrm{STYLERANK}(\mathcal{P}_{\text{accessory}}, \mathcal{D}_{\text{accessory}}, \mathcal{A}_{p})
\]

\[
\text{outfit} =
\{ \text{top}, \text{match}_{\text{top-bottom}}, \text{match}_{\text{top-accessory}}, \dots \}
\]
\end{center}

\begin{algorithm}
\caption{STYLERANK}
\label{alg1}
\begin{algorithmic}[1]
\State \textbf{Input:} Products $P$, Distance function $D$, Appearance count $A$, Control parameter $\lambda$
\State Rank products in $P$ based on ascending $D$
\For{each  product $p \in P$}
    \State Assign rank $R_D(p) = rank$
\EndFor
\State Rank products in $P$ based on descending $A$
\For{each  product $p \in P$}
    \State Assign rank $R_A(p) = rank$
\EndFor
\For{each  product $p \in P$}
    \State $S(p) \gets R_D(p) + \lambda \cdot R_A(p)$
\EndFor
\State \textbf{Output:} Select min($S(P)$)
\end{algorithmic}
\end{algorithm}

\section{Experimental Evaluation}
\label{sec:experimental}

This section presents a comprehensive evaluation of our proposed methods. We utilize both qualitative and quantitative metrics across different datasets to assess the performance of each module in isolation. Additionally, we provide a detailed analysis to justify the design choices made throughout our approach.

For the attribution system, we utilized approximately 15,000 fully annotated products, where domain experts meticulously categorized each garment according to its full set of attributes, including division (e.g., Tops, Bottoms, Accessories), secondary division, specific category (e.g., T-shirts, Jeans, Bags), material, fabric characteristics, color, fit, design detailing, aesthetic and many more.

Next for the virtual try-off task, we used two datasets; the publicly available VITON-HD \cite{vton-hd} and a proprietary dataset obtained from a major online fashion shopping platform. These datasets serve as robust benchmarks for the virtual try-off task due to their comprehensive coverage of diverse garment types and real-world use cases. Specifically, the VITON-HD dataset comprises 13,679 high-resolution (1024 × 768) image pairs featuring frontal half-body models and their corresponding upper-body garments. Originally curated for the Virtual Try-On (VTON) task, this dataset is well-suited for our study as it provides the necessary image pairs (C, HC), where C represents the garment in a lay-down view (Figure \ref{fig:garment}a), and HC is the human-worn garment image (Figure \ref{fig:garment}b). The dataset sourced from the major online fashion shopping platform consists of 24,568 images spanning a diverse range of garment categories. Specifically, it includes 12,771 images of upper-body garments (e.g., T-shirts, shirts, blouses), 7,158 images of bottoms (e.g., pants, skirts), and 4,639 images of dresses. For both datasets, we generated the garment masks using the SegFormer \cite{DBLP:journals/corr/abs-2105-15203} model for garment segmentation.

To train the recommendation system, we leveraged a dataset of 25,000 distinct outfits. Of these, 15,000 outfits were curated by domain fashion experts, ensuring they represent well-structured and aesthetically coherent ensembles. The remaining 10,000 outfits were generated using our recommendation system and subsequently reviewed and approved by the same team of domain experts, guaranteeing their stylistic quality. This hybrid approach allowed us to scale the dataset while maintaining high-quality outfit compositions, leading to more effective training and evaluation of our recommendation model.

The experiments were conducted on a server with 4 NVIDIA A10G GPUs. The model training takes approximately 8 hours for the virtual try-off model, while less than 30 minutes were required for the product attribution and outfit recommendation models.

\subsection{Metrics}
\label{sec:metrics}

For the attribution system, we employ accuracy as the primary evaluation metric, which is the standard for classification tasks. To ensure a holistic assessment, we compute the average accuracy across all attributes, as well as each attribute individually.

For the virtual try-off task, we utilize five widely adopted metrics to assess the similarity between synthesized images and real images; Structural Similarity Index (SSIM) \cite{SSIM}, Fréchet Inception Distance (FID) \cite{Seitzer2020FID}, Kernel Inception Distance (KID) \cite{bińkowski2021demystifyingmmdgans}, Learned Perceptual Image Patch Similarity (LPIPS) \cite{LPIPS}, and DISTS \cite{DISTS}. A comprehensive analysis conducted by TryOffDiff \cite{velioglu2024tryoffdiffvirtualtryoffhighfidelitygarment} demonstrated that DISTS is particularly well-suited for the virtual try-off task, as it effectively captures visual fidelity and fine-grained details in synthesized images. Consequently, we prioritize DISTS in evaluating the performance of our models.

To assess the effectiveness of the outfit recommendations generated by our algorithm, we conducted a qualitative evaluation with a team of domain experts specializing in fashion styling and aesthetics. The experts were presented with 10,000 outfits and asked to evaluate them based on two key criteria: stylistic coherence and variety. For stylistic coherence, experts were instructed to approve an outfit if all components were appropriately matched and formed a cohesive, aesthetically pleasing ensemble, or to reject the outfit if any individual item was deemed incompatible in terms of fashion styling. In addition to evaluating individual outfits, the experts assessed the generated variety by rejecting outfits when specific items were frequently selected across multiple recommendations, indicating a lack of diversity.

\subsection{Attribution System Efficiency}

The proposed attribution system demonstrates exceptional performance by effectively leveraging multi-modal information for detailed product categorization. To evaluate its efficiency, we test 3 configurations: (1) use FashionCLIP as a classifier and prompts to retrieve the desired attribute, (2) combine ResNet-152 for visual features with an NLP model, called Mxbai \cite{li2023angle} for title and description processing, and (3) utilize FashionCLIP which integrates both image and textual attributes in a unified multi-modal framework.

Table~\ref{tab:attribution_accuracy} shows the results of the product attribution evaluation across all attributes. While the FashionCLIP is trained on cloth images and has knowledge about attributes such as division (e.g., Top, Bottom, etc.) or color, it performs poorly on more detailed attributes like hem type or pocket type. Overall, its performance is very low when considering all possible attributes of interest to this system. Combining visual and textual from different feature extractors and the train a classification head on them performs well. However, the highest performance was achieved using FashionCLIP \cite{Chia2022} as a feature extractor and a custom classification head, which fully exploits the synergy between visual and textual data through a shared embedding space. In addition to overall improvements, the attribution system excelled in the critical domain of division classification (e.g., Tops, Bottoms, Outerwear), achieving an impressive accuracy of 99.6\%.

\begin{table}[h!]

  \caption{Product Attribution Accuracy}
\label{tab:attribution_accuracy}
  \begin{tabular}{cc}
    \toprule
\textbf{Methodology} & \textbf{Accuracy (\%)} \\
    \midrule
FashionCLIP \cite{Chia2022} & 73.8 \\
ResNet \cite{ResNet} + Mxbai\cite{li2023angle} + Classifier  & 89.1 \\ 
FashionCLIP + Classifier (Proposed System) & \textbf{92.3} \\
  \bottomrule
\end{tabular}
\end{table}

\subsection{Virtual Try-Off System Performance}

In Table \ref{table:quantitative}, we provide a detailed quantitative comparison of our proposed virtual try-off system against TileGAN \cite{article_TileGAN}, TryOffDiff \cite{velioglu2024tryoffdiffvirtualtryoffhighfidelitygarment}, and the recently introduced Multi-Garment TryOffDiff (MGT) \cite{velioglu2025mgtextendingvirtualtryoff}, using the VITON-HD dataset \cite{vton-hd}. While MGT proposes a unified diffusion-based architecture to support multi-garment try-off and claims generalization across domains—including from DressCode to VITON-HD, their reported numbers contradict this claim. As shown in their evaluation, MGT underperforms not only our method across nearly all key perceptual and structural metrics, but also falls short of the original single-garment TryOffDiff baseline on several core indicators, most notably SSIM, LPIPS, and DISTS. Specifically, MGT reports lower SSIM (78.1 vs. 79.5), worse LPIPS (36.3 vs. 32.4), and inferior DISTS (24.7 vs. 23.0) compared to TryOffDiff, highlighting the model’s limitations in cross-dataset generalization despite its architectural enhancements.

Our model, by contrast, is explicitly designed and tuned for robustness across diverse garment types and body poses, handling both full-body and half-body images in production without requiring garment-specific segmentation or category-level tuning. It achieves state-of-the-art performance on DISTS, widely regarded as the most appropriate for evaluating garment fidelity and visual coherence in virtual try-off systems \cite{velioglu2024tryoffdiffvirtualtryoffhighfidelitygarment} and delivers consistent improvements on LPIPS and KID, confirming its advantage in generating photorealistic and perceptually accurate results. Furthermore, unlike TEMU-VTOFF \cite{lobba2025inversevirtualtryongenerating}, whose resource-heavy architecture is impractical for deployment on consumer-grade GPUs (<24GB VRAM), our model is fully optimized for efficient real-world usage. 

\begin{figure}
\centering
\includegraphics[width=\linewidth]{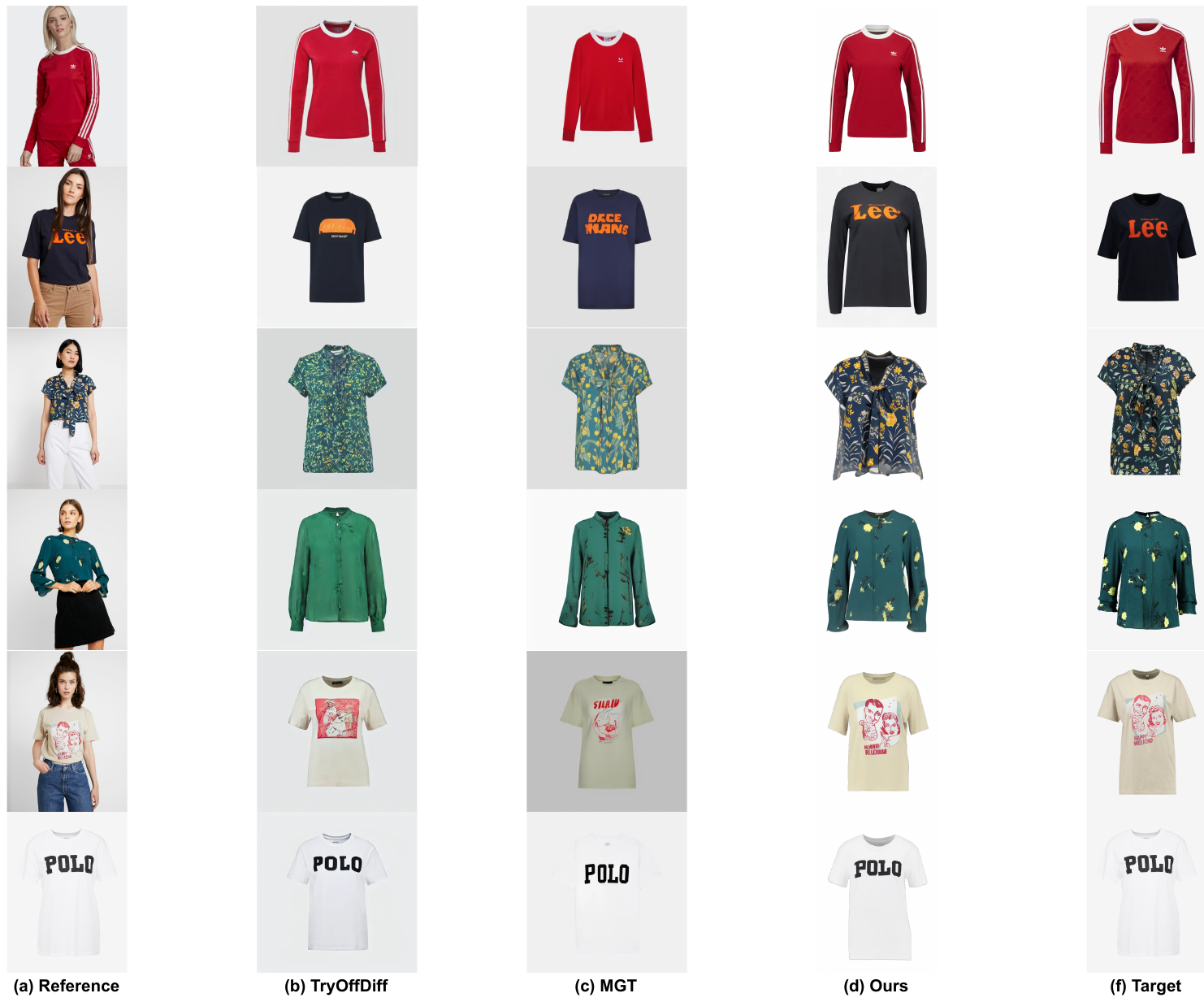}
\caption{Comparison against TryOffDiff \cite{velioglu2024tryoffdiffvirtualtryoffhighfidelitygarment} and MGT \cite{velioglu2025mgtextendingvirtualtryoff}}
\label{fig:qualitative_comp}
\end{figure}

In Figure  \ref{fig:qualitative_comp}, we present the results of a qualitative comparison. As TileGan has not publicly released its code or pretrained models we conduct the comparison against TryOffDiff and MGT. Our method demonstrates a clear and consistent advantage in terms of both texture fidelity and pattern accuracy against TryOffDiff and MGT, setting a new benchmark for the virtual try-off task. More virtual try-off example outputs can be found in Section \ref{appendix}.

\begin{table}[h!]
  \caption{Virtual Try-Off Performance}
\label{table:quantitative}
  \begin{tabular}{cccccc}
    \toprule
    \textbf{Method} & \textbf{SSIM ↑} & \textbf{FID ↓} & \textbf{KID ↓} & \textbf{LPIPS ↓} & \textbf{DISTS ↓} \\ 
    \midrule
TileGan \cite{article_TileGAN} & 70.96 & 39.8 & N/A & N/A & N/A \\ 
TryOffDiff \cite{velioglu2024tryoffdiffvirtualtryoffhighfidelitygarment} & \textbf{79.5} & 25.1 & 8.9 & 32.4 & 23.0 \\ 
MGT \cite{velioglu2025mgtextendingvirtualtryoff} & 78.1 & \textbf{21.9} & 8.9 & 36.3 & 24.7 \\ 
Ours & 71.99 & 25.3 & \textbf{2.01} & \textbf{17.22} & \textbf{21.01} \\ 
  \bottomrule
\end{tabular}
\end{table}

\subsection{Recommendation System Performance}

The main objective of this work lies in the development of a robust outfit recommendation system capable of creating high-styling outfits that drive customer engagement and purchases. Both the attribution and virtual try-off methods are integral components of this pipeline, directly influencing the recommendation system's performance. The proposed recommendation system is highly dependent on the precise attribution of garment features and the realistic and high-quality representation of garment images.

Table \ref{table:recommendation_comparison} compares the accuracy of our outfit recommendation system with the baseline OutfitTransformer \cite{Sarkar_2023_WACV} method, based on evaluations by human experts both in terms of outfit aesthetic compatibility as well as outfit diversity. Our method, which incorporates lay-down images and meaningful product attributes, significantly outperforms both the OutfitTransformer baseline and the approach using raw images and textual descriptions. 

While OutfitTransformer achieves competitive accuracy when diversity is not considered, our evaluation emphasizes the importance of variety in recommendations, a factor where our system excels due to the use of the STYLERANK algorithm. The results also highlight that relying solely on raw images and textual descriptions (titles and descriptions) does not allow the system to reach optimal efficiency. Instead, the highest accuracy (91.3\%) is obtained when leveraging lay-down images and structured attribute information, demonstrating the necessity of the preprocessing pipeline we have developed. This underscores the importance of our prior efforts in structuring and enriching the data, ultimately enabling a more effective and context-aware recommendation system. We provide samples of outfits that are currently displayed on a major online shopping store in Section \ref{appendix}.

\begin{table}[H]

  \caption{Outfit Recommendation Accuracy}
\label{table:recommendation_comparison}
  \begin{tabular}{cc}
    \toprule
\textbf{Input Type} & \textbf{Accuracy (\%)} \\
    \midrule
Ours without STYLERANK & 55.7\\
OutfitTransformer \cite{Sarkar_2023_WACV} & 75.2 \\ 
Ours (Raw Images + Title + Description) & 79.8 \\ 
Ours (Lay-down Images + Attributes) & \textbf{91.3} \\
  \bottomrule
\end{tabular}
\end{table}

\subsection{Deployed System Impact}
To assess the real-world impact of our fully deployed fashion e-commerce system, we conducted a three-month A/B testing campaign across the major retailer platform. The results demonstrate clear business and user value driven primarily by the recommendation engine, in combination with the attribution and try-off components. Key monetization metrics improved significantly: Revenue Per Visit (RPV) increased by 6.60\%, while Net Margin Per Visit (NMPV)—which accounts for both product and shipping margins—rose by 6.62\%, indicating that our system not only boosts sales volume but does so in a profit-conscious manner. Similarly, Conversion Rate (CVR), defined as the ratio of purchases to visits, saw a 5.61\% uplift, reflecting stronger relevance and engagement from users exposed to our recommendations. These improvements are particularly meaningful given the scale of the platform and the relative stability of the Product Margin Rate (PMR), which remained essentially flat at -0.11\%, suggesting that profit gains were not offset by increased discounting or lower-margin sales.

Notably, the Click-Through Rate (CTR)—measured as the ratio of product clicks to impressions—also showed a significant lift of approximately 8.2\%, consistent with the observed improvements in CVR and RPV. This confirms that users not only saw more relevant content but were also more compelled to explore and ultimately convert. Overall, these results underscore the commercial viability of our system and its ability to simultaneously enhance user satisfaction and core business outcomes.
\section{Conclusions}
\label{sec:conc}

We presented TrendGen, an advanced AI-powered outfit recommendation and display system that enhances online fashion shopping. It combines FashionCLIP-based multi-modal embeddings, a latent diffusion model for virtual try-off, and a TripletNet-based recommendation engine to deliver trend-aware, high-quality outfit suggestions. Evaluations on both proprietary and public datasets show TrendGen outperforms existing methods in product attribution, virtual try-off realism, and outfit aesthetics.

There are several exciting opportunities for improving TrendGen by incorporating real-time personalization based on user behavior, integrating virtual try-ons to better simulate in-store experiences, and using background image generation to enhance outfit presentation and engagement.

\bibliographystyle{ACM-Reference-Format}
\bibliography{sample-base}

\appendix

\section{TrendGen: More Qualitative Results}
\label{appendix}
\begin{figure*}
    \centering
    
    \begin{subfigure}{\linewidth}
        \centering
        \adjustbox{max width=0.75\linewidth}{\includegraphics{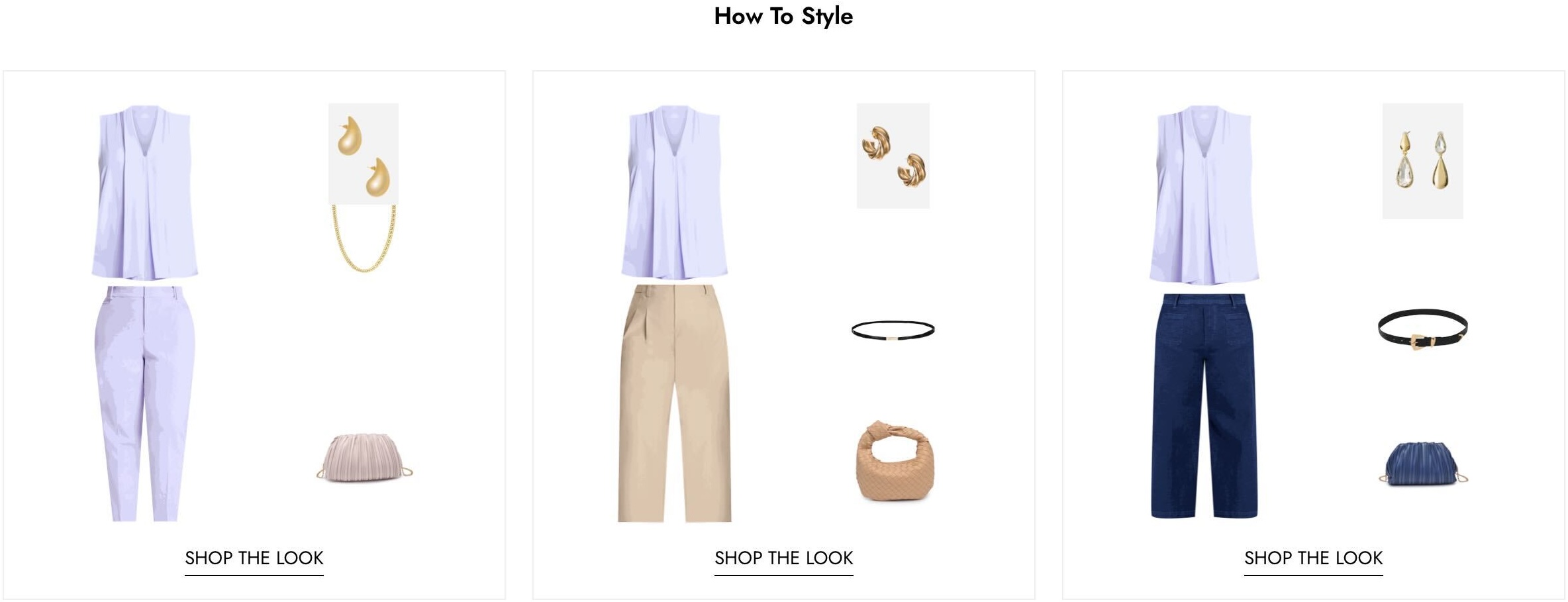}}
    \end{subfigure}
    
    \vspace{10pt}
    \begin{subfigure}{\linewidth}
        \centering
        \adjustbox{max width=0.75\linewidth}{\includegraphics{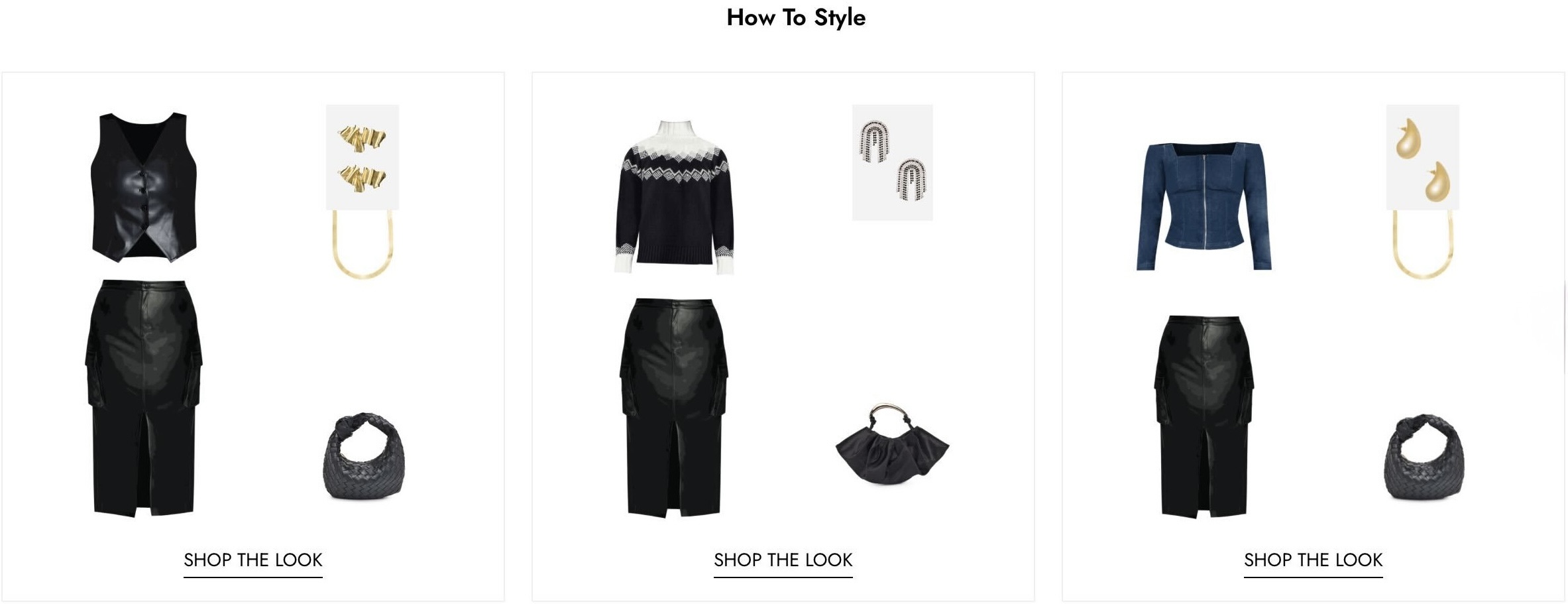}}
    \end{subfigure}

    \vspace{10pt}
    \begin{subfigure}{\linewidth}
        \centering
        \adjustbox{max width=0.75\linewidth}{\includegraphics{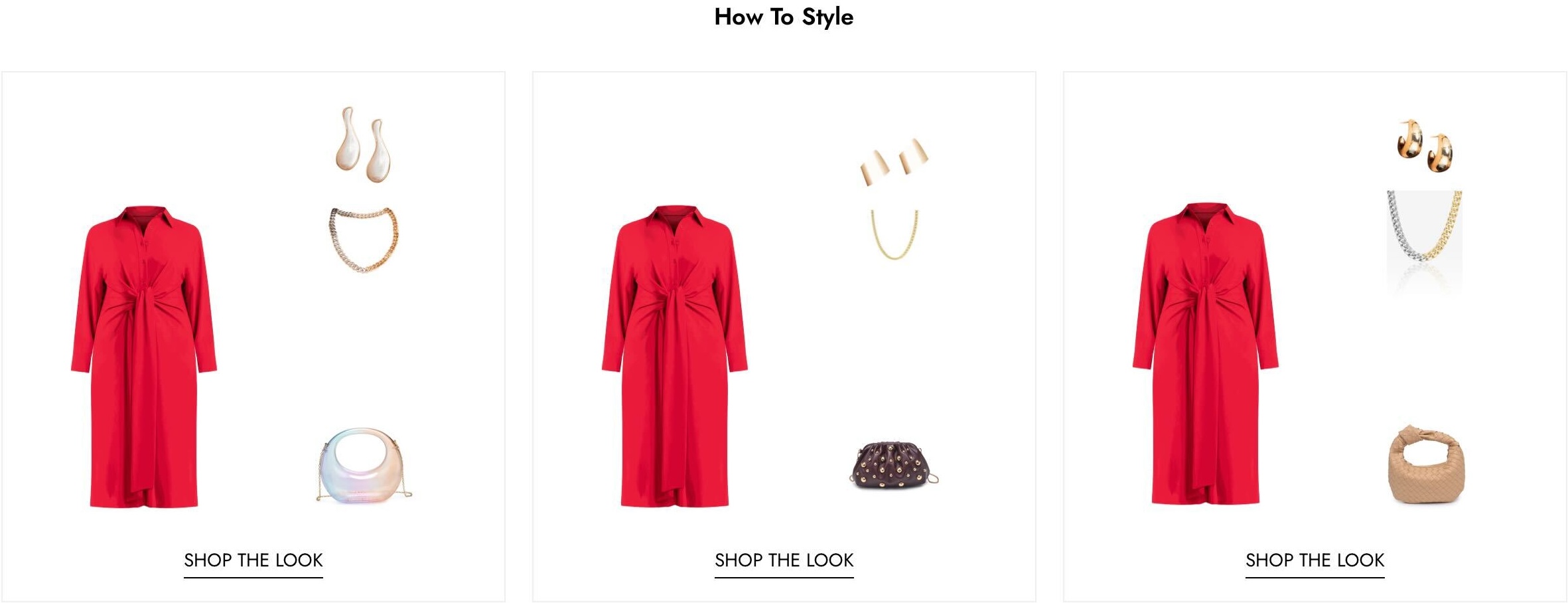}}
    \end{subfigure}

    \vspace{10pt}
    \begin{subfigure}{\linewidth}
        \centering
        \adjustbox{max width=0.75\linewidth}{\includegraphics{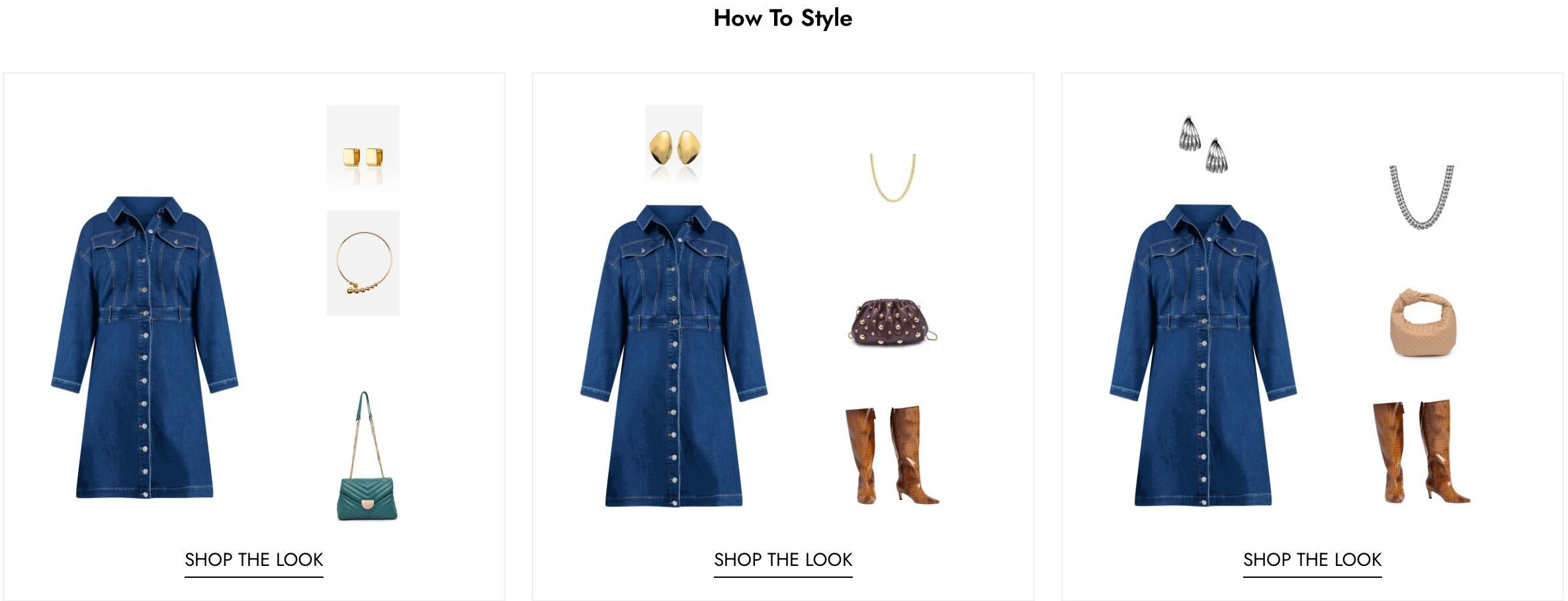}}
    \end{subfigure}

    \caption{Three distinct styling approaches, each representing different fashion aesthetics. These are visually connected through color coordination, accessories, and outfit choices.}
    \label{fig:styling_series}

\end{figure*}

\begin{figure*}
\centering
\includegraphics[width=\linewidth]{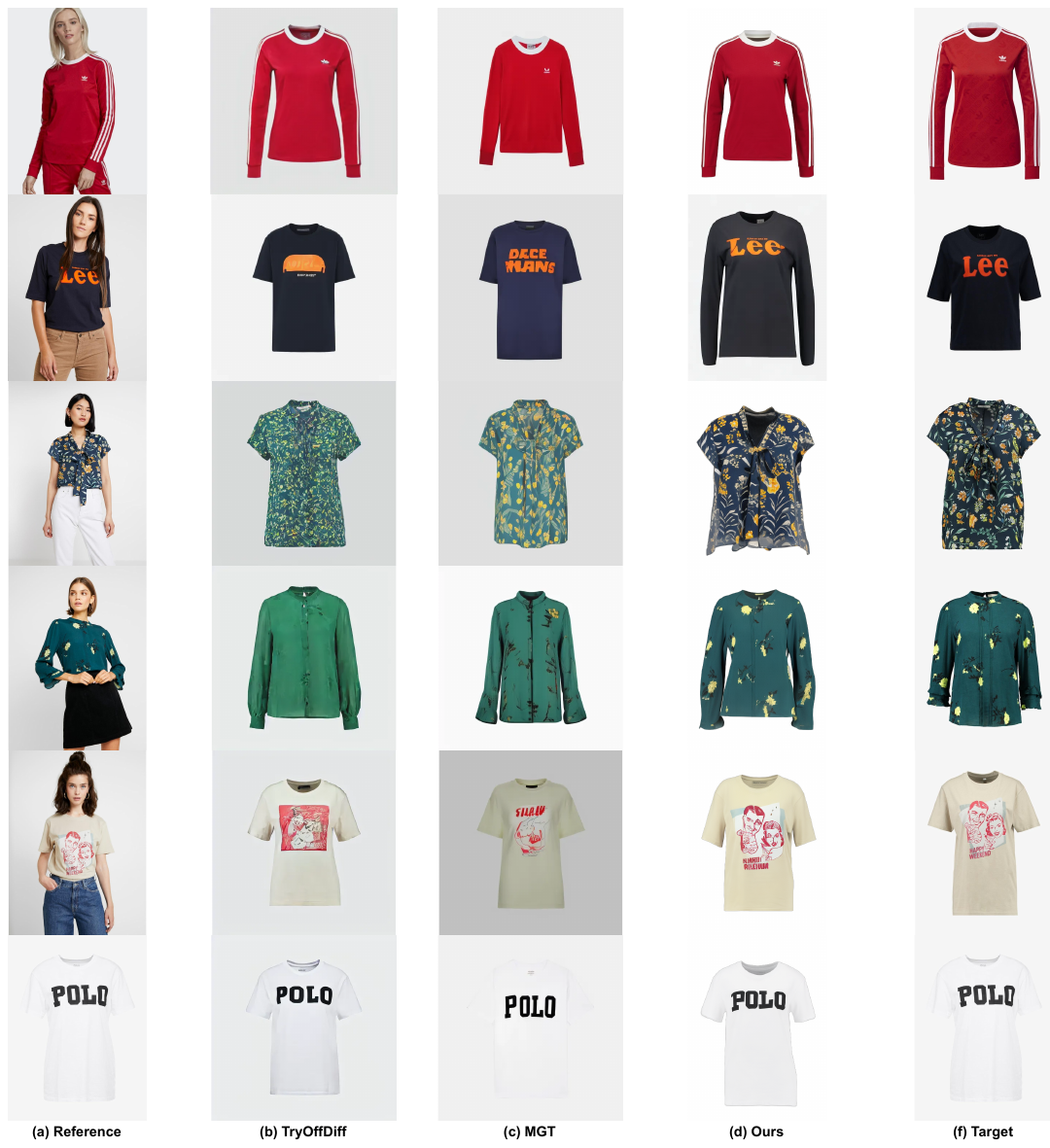}
\caption{Qualitative comparison against TryOffDiff \cite{velioglu2024tryoffdiffvirtualtryoffhighfidelitygarment} and MGT \cite{velioglu2025mgtextendingvirtualtryoff}}
\label{fig:qual_comp}
\end{figure*}

\begin{figure*}
\centering
\includegraphics[width=\linewidth]{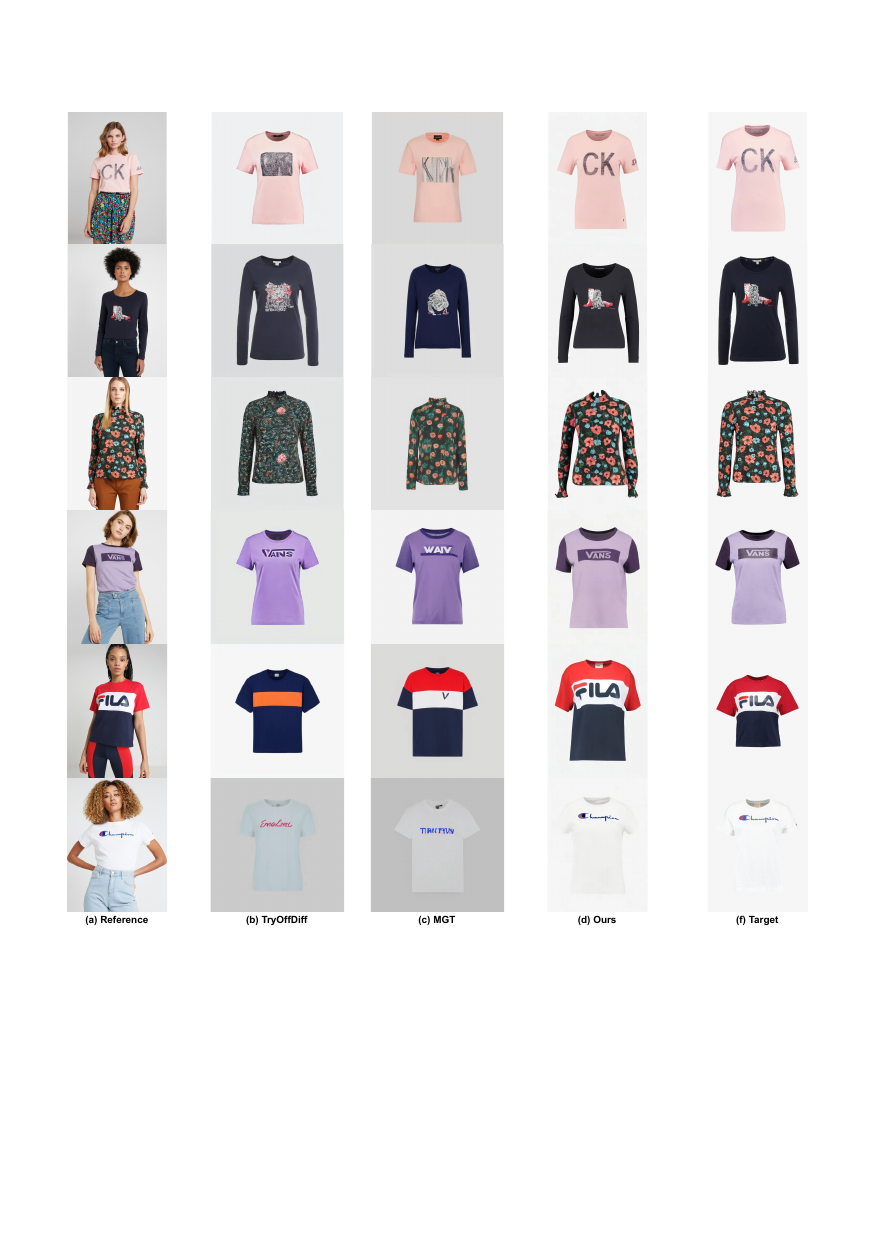}
\caption{Qualitative comparison against TryOffDiff \cite{velioglu2024tryoffdiffvirtualtryoffhighfidelitygarment} and MGT \cite{velioglu2025mgtextendingvirtualtryoff}}
\label{fig:qual_comp1}
\end{figure*}

\begin{figure*}
\centering
\includegraphics[width=\linewidth]{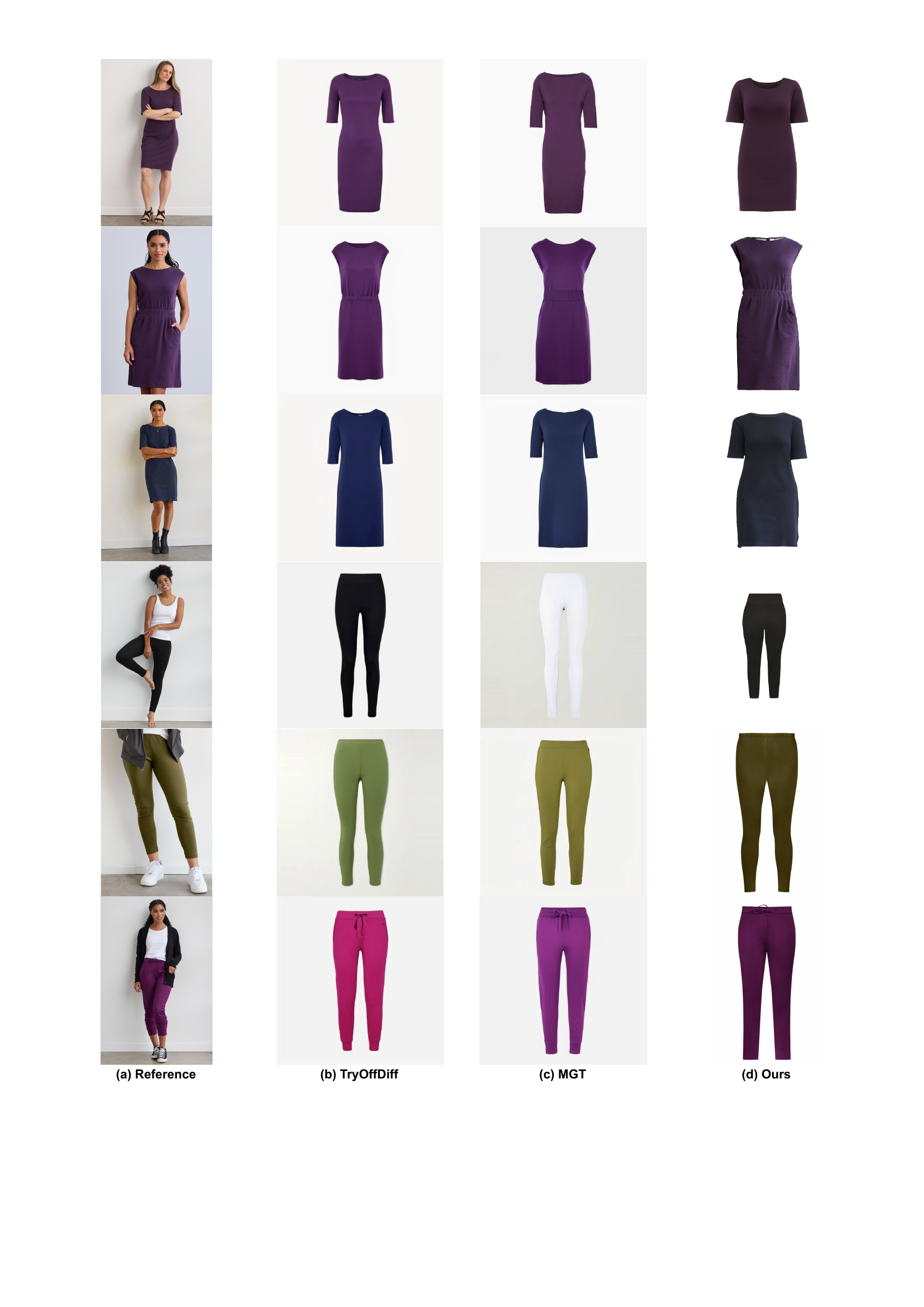}
\caption{Qualitative comparison against TryOffDiff \cite{velioglu2024tryoffdiffvirtualtryoffhighfidelitygarment} and MGT \cite{velioglu2025mgtextendingvirtualtryoff}}
\label{fig:fbb}
\end{figure*}

Figure \ref{fig:styling_series} shows some of the outfits that were generated by our outfit recommender and are currently displayed in a major online shopping store. In Figures \ref{fig:qual_comp} and \ref{fig:qual_comp1} we showcase some more comparisons against state-of-the-art TryOffDiff \cite{velioglu2024tryoffdiffvirtualtryoffhighfidelitygarment}. In Figure \ref{fig:fbb}, we display additional outputs of our Virtual Try-Off component for both half-body and full-body inputs from our dataset. 

\section{STYLERANK: $\lambda$ parameter effect }
\label{appendix_stylerank}

To evaluate the impact of the control parameter $\lambda$ in the STYLERANK algorithm (Algorithm~\ref{alg1}), we perform an ablation study by varying $\lambda \in \{0, 0.25, 0.5, 1, 1.5, 2, 3\}$. This parameter determines the trade-off between aesthetic compatibility and product repeatability in the final ranking of candidate items.

As defined in Algorithm~\ref{alg1}, the final score \( S(p) \) for each product \( p \) is computed as:
\[
S(p) = R_D(p) + \lambda \cdot R_A(p)
\]
where \( R_D(p) \) is the rank based on visual compatibility (from kNN on TripletNet embeddings), and \( R_A(p) \) is the rank based on historical appearance frequency. A lower \( S(p) \) indicates a better match.

Setting \(\lambda = 0\) ignores product frequency entirely and ranks items based solely on compatibility. Conversely, higher \(\lambda\) values place more emphasis on frequency, which sacrifices product compatibility for diversity.

Table~\ref{tab:lambda_ablation} summarizes the outfit approval rate by expert evaluators across different values of $\lambda$. The best result is achieved at $\lambda = 1$, indicating a well-balanced integration of both compatibility and frequency.

\begin{table}[h]
\centering
\caption{Ablation study on $\lambda$ in STYLERANK. Performance is reported as expert-verified outfit approval rate (\%).}
\label{tab:lambda_ablation}
\begin{tabular}{c|ccccccc}
\toprule
$\lambda$ & 0 & 0.25 & 0.5 & 1.0 & 1.5 & 2.0 & 3.0 \\
\midrule
Accuracy (\%) & 55.7 & 78.4 & 89.7 & \textbf{91.3} & 87.1 & 85.9 & 77.2 \\
\bottomrule
\end{tabular}
\end{table}

These results show that while purely compatibility-based ranking (\(\lambda = 0\)) severely underperforms, as diversity is an important factor of the expert's approval, moderate incorporation of frequency signals improves both diversity and outfit cohesion. Overweighting appearance count (\(\lambda > 1\)), however, leads to an increasing performance decline, due to reduced stylistic compatibility across recommendations.

\end{document}